\pdfoutput=1
\documentclass{ecai}
\usepackage{times}
\usepackage{graphicx}
\usepackage{latexsym}

\ecaisubmission   % inserts page numbers. Use only for submission of paper.
\newif\ifsubmission
\submissiontrue

\newif\ifanonymous
\anonymousfalse
\newif\iffull
\fullfalse
\newcommand\CHANGED[1]{\ifsubmission #1\else {\color{red} #1}\fi}

\usepackage{hyperref}

\expandafter\def\expandafter\UrlBreaks\expandafter{\UrlBreaks%  save the current one 
	\do\a\do\b\do\c\do\d\do\e\do\f\do\g\do\h\do\i\do\j% 
	\do\k\do\l\do\m\do\n\do\o\do\p\do\q\do\r\do\s\do\t% 
	\do\u\do\v\do\w\do\x\do\y\do\z\do\A\do\B\do\C\do\D% 
	\do\E\do\F\do\G\do\H\do\I\do\J\do\K\do\L\do\M\do\N% 
	\do\O\do\P\do\Q\do\R\do\S\do\T\do\U\do\V\do\W\do\X% 
	\do\Y\do\Z}

\usepackage{amssymb}

\usepackage{amsmath}
\usepackage{accents}
\usepackage{bm}
\usepackage{csquotes}
\usepackage[binary-units=true, per-mode=symbol, per-symbol=/]{siunitx}
\usepackage{numprint}
\npdecimalsign{.}
\npthousandsep{,}
\usepackage{booktabs}
\usepackage{multirow}
\usepackage{hhline}

\usepackage{tikz}
\usetikzlibrary{positioning}
\usetikzlibrary{backgrounds}
\usepackage[draft]{tikzpeople}

\makeatletter
\tikzset{
    database/.style={
        path picture={
            \draw (0, 1.5*\database@segmentheight) circle [x radius=\database@radius,y radius=\database@aspectratio*\database@radius];
            \draw (-\database@radius, 0.5*\database@segmentheight) arc [start angle=180,end angle=360,x radius=\database@radius, y radius=\database@aspectratio*\database@radius];
            \draw (-\database@radius,-0.5*\database@segmentheight) arc [start angle=180,end angle=360,x radius=\database@radius, y radius=\database@aspectratio*\database@radius];
            \draw (-\database@radius,1.5*\database@segmentheight) -- ++(0,-3*\database@segmentheight) arc [start angle=180,end angle=360,x radius=\database@radius, y radius=\database@aspectratio*\database@radius] -- ++(0,3*\database@segmentheight);
        },
        minimum width=2*\database@radius + \pgflinewidth,
        minimum height=3*\database@segmentheight + 2*\database@aspectratio*\database@radius + \pgflinewidth,
    },
    database segment height/.store in=\database@segmentheight,
    database radius/.store in=\database@radius,
    database aspect ratio/.store in=\database@aspectratio,
    database segment height=0.1cm,
    database radius=0.25cm,
    database aspect ratio=0.35,
}
\makeatother

\DeclareMathSymbol{\widetildesym}{\mathord}{largesymbols}{"65}
    \newcommand\lowerwidetildesym{%
      \text{\smash{\raisebox{-1.3ex}{%
        $\widetildesym$}}}}
            \newcommand\fixwidetilde[1]{%
              \mathchoice
                {\accentset{\displaystyle\lowerwidetildesym}{#1}}
        {\accentset{\textstyle\lowerwidetildesym}{#1}}
        {\accentset{\scriptstyle\lowerwidetildesym}{#1}}
        {\accentset{\scriptscriptstyle\lowerwidetildesym}{#1}}
    }

\let\oldparagraph=\paragraph
\renewcommand\paragraph[1]{\oldparagraph{#1.}}

\newcommand{\X}{\ensuremath{\mathbf{X}}}

\newcommand{\region}{\ensuremath{\mathbf{R}}}
\newcommand{\partition}{\ensuremath{\mathcal{P}}}

\newcommand{\coolname}{Crypto\-SPN}

\newcommand{\papertitle}{\coolname: Privacy-preserving Sum-Product Network Inference}

\begin{document}

\title{\papertitle}

\author{Amos Treiber\institute{Cryptography and Privacy Engineering Group, TU Darmstadt,
Germany, email addresses: \{treiber,weinert,schneider\}@encrypto.cs.tu-darmstadt.de} \and Alejandro Molina$^2$ \and Christian Weinert$^1$\and \\ Thomas Schneider$^1$ \and Kristian Kersting\institute{Artificial Intelligence and Machine Learning Lab, TU Darmstadt, Germany, email addresses: \{molina,kersting\}@cs.tu-darmstadt.de}
}

\maketitle
\bibliographystyle{ecai}

\begin{abstract}
    AI algorithms, and machine learning~(ML) techniques in particular, are increasingly important to individuals' lives, but have caused a range of privacy concerns addressed by,~e.g., the~European GDPR. 
    Using cryptographic techniques, it is possible to perform inference tasks remotely on sensitive client data in a privacy-preserving way: the server learns nothing about the input data and the model predictions, while the client learns nothing about the~ML model~(which is often considered intellectual property and might contain traces of sensitive data).
    While such privacy-preserving solutions are relatively efficient, they are mostly targeted at neural networks, can degrade the predictive accuracy, and usually reveal the network's topology. 
    Furthermore, existing solutions are not readily accessible to~ML experts, as prototype implementations are not well-integrated into~ML frameworks and require extensive cryptographic knowledge.

    In this paper, we present \coolname{}, a framework for privacy-preserving inference of sum-product networks~(SPNs). SPNs are a tractable probabilistic graphical model that allows a range of exact inference queries in linear time. Specifically, we show how to efficiently perform~SPN inference via secure multi-party computation~(SMPC) without accuracy degradation while hiding sensitive client and training information with provable security guarantees.
    Next to foundations, \coolname{} encompasses tools 
    to easily transform existing~SPNs into privacy-preserving executables.
    Our empirical results demonstrate that \coolname{} achieves highly efficient and accurate inference in the order of seconds for medium-sized SPNs.
\end{abstract}

\section{INTRODUCTION}
\label{sec_intro}
In our increasingly connected world, the abundance of user information and availability of data analysis techniques originating from artificial intelligence~(AI) research has brought machine learning~(ML) techniques into daily life.
While these techniques are already deployed in many applications like credit scoring, medical diagnosis, biometric verification, recommender systems, fraud detection, and language processing, emerging technologies such as self-driving cars will further increase their popularity.

\paragraph{Privacy Concerns for ML Applications}
These examples show that progress in AI research certainly has improved user experience, potentially even saving lives when employed for medical or safety purposes.
However, as the prevalent usage in modern applications often requires the processing of massive amounts of sensitive information, the impact on user privacy has come into the public spotlight.

This culminated in 
privacy regulations such as the European General Data Protection Regulation~(GDPR), which came into effect in~2018.
Not only does the~GDPR provide certain requirements to protect user data, it also includes restrictions on decisions based on user data, which may be interpreted as a~\enquote{right to explanation}~\cite{goodman_european_2017}. 
Luckily, new deep probabilistic models that encode the joint distribution, such as sum-product networks~(SPNs)~\cite{Poon2011}, can indicate whether the model is fit to predict the data at hand, or raise a warning otherwise.
This increases trust as they~\enquote{know when they do not know}. 
Moreover, SPNs can also perform inference with missing information~\cite{peharz2019random}, an important aspect for real-life applications. 

Generally, probabilistic graphical models~\cite{Koller2009} provide a framework for understanding what inference and learning are, and have therefore emerged as one of the principal theoretical and practical approaches to~ML and~AI~\cite{ghahramani_2015}.
However, one of the main challenges in probabilistic modeling is the trade-off between the expressivity of the models and the complexity of performing various types of inference, as well as learning them from data.
This inherent trade-off is clearly visible in powerful -- but intractable -- models like~Markov random fields, (restricted)~Boltzmann machines, (hierarchical)~Dirichlet processes, and variational autoencoders. 
Despite these models' successes, performing inference on them resorts to approximate routines.
Moreover, learning such models from data is generally harder as inference is a sub-routine of learning, requiring simplified assumptions or further approximations.
Having guarantees on tractability at inference and learning time is therefore a highly desired property in many real-world scenarios. 

Tractable graphical models such as~SPNs guarantee exactly this: performing exact inference for a range of queries.
They compile probabilistic inference routines into efficient computational graphs similar to deep neural networks, but encode a joint probability distribution.
As a result, they can not only be used for one~ML task, but support many different tasks by design, ranging from outlier detection~(joint distribution) to classification or regression~(conditional inference).
They have been successfully used in numerous real-world applications such as image classification, completion and generation, scene understanding, activity recognition, and language and speech modeling. 
Despite these successes, it is unclear how one can develop an SPN framework that is GDPR-friendly.

\begin{figure}[t]
    \centering
    \resizebox{\columnwidth}{!}{%
    \begin{tikzpicture}[yscale = 0.8]
        \node [alice, minimum height=1.2cm, minimum width=0.5cm] (alice) at (-4, 0) {};
        \node [database, database radius=0.5cm, database segment height=0.25cm] (server) at (4,0) {};
        \node [] (model) at (5.00,0) {$P$};
        
        \draw [->, shorten <=0.25cm, shorten >=0.25cm] ([yshift=0.25cm]alice.east) -- ([yshift=0.25cm]server.west) node[pos=0.5, yshift=0.25cm] {$\bm{X}$};
        \draw [->, shorten <=0.25cm, shorten >=0.25cm] ([yshift=-0.25cm]server.west) -- ([yshift=-0.25cm]alice.east) node[pos=0.5, yshift=-0.25cm] {$P(\bm{X}) = 0.9$};
        
        \draw (-4.5,-1.5) -- (5.0,-1.5) node[above,xshift=-8.125cm] {Conventional MLaaS} node[below,xshift=-8.125cm] {Private ML inference};
        
        \node [alice, minimum height=1.2cm, minimum width=0.5cm] (alice) at (-4, -3.25) {};
        \node [] (clientL) at (-3.9875,-4.325) {Client};
        \node [database, database radius=0.5cm, database segment height=0.25cm] (server) at (4,-3.25) {};
        \node [] (serverL) at (3.9875,-4.325) {Server};
        
        \node [draw, thick, dashed, rectangle, minimum height=1.625cm, minimum width=4.5cm] (box) at (0,-3.25) {};
        
        \draw [shorten <=0.25cm, shorten >=5.75cm] ([yshift=0.25cm]alice.east) -- ([yshift=0.25cm]server.west) node[pos=0.5, xshift=-2.75cm, yshift=0.25cm] {$\bm{X}$};
        \draw [->, shorten <=5.75cm, shorten >=0.25cm] ([yshift=-0.25cm]server.west) -- ([yshift=-0.25cm]alice.east) node[pos=0.5, xshift=-2.75cm, yshift=-0.25cm] {$0.9$};
        
        \draw [shorten <=0.25cm, shorten >=5.8cm] (server.west) -- (alice.east) node[pos=0.5, xshift=2.8cm, yshift=0.25cm] {$P$};
        
        \draw [->, shorten <=1.25cm, shorten >=4.5cm, dashed] ([yshift=0.25cm]alice.east) -- ([yshift=0.25cm]server.west) node[pos=0.5, xshift=-1.625cm, yshift=0.25cm] {$\fixwidetilde{\bm{X}}$};
        \draw [shorten <=1.25cm, shorten >=4.5cm, dashed] ([yshift=-0.25cm]alice.east) -- ([yshift=-0.25cm]server.west) node[pos=0.5, xshift=-1.6cm, yshift=-0.3cm] {$\fixwidetilde{P}(\fixwidetilde{\bm{X}})$};
        
        \draw [->, shorten <=1.25cm, shorten >=4.5cm, dashed] (server.west) -- (alice.east) node[pos=0.5, xshift=1.625cm, yshift=0.25cm] {$\fixwidetilde{P}$};
        
        \node [draw, rectangle, minimum height=1.125cm, minimum width=1.625cm] (innerbox) at (0,-3.25) {SMPC};
    \end{tikzpicture}
    }
    \vspace{-2em}
    \caption{\emph{Conventional} MLaaS~(top) vs.~\emph{private}~ML inference~(bottom) on client feature vector~$\bm{X}$ using the server's model~$P$. In private~ML inference, the input of both client and server is protected with a cryptographic~SMPC protocol: the parties learn only encrypted values~$\tilde{\bm{X}}$ and~$\tilde{P}$, respectively.}
    \label{fig:ml_dummies}
\end{figure}
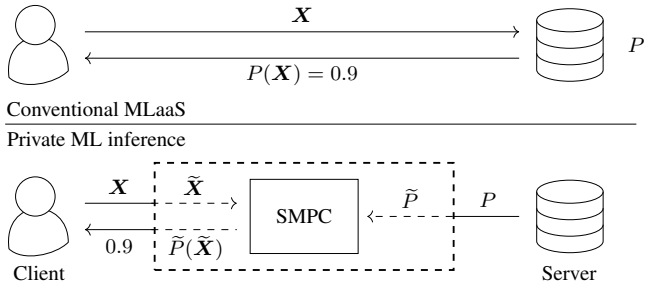

As a naive solution, SPN tasks can be performed only on client devices to ensure that no sensitive information is handed out, but this requires the service provider to ship a trained model to clients, thereby giving up valuable intellectual property and potentially leaking sensitive data as such models often contain traces of sensitive training data, e.g., due to unintended memorization~\cite{memorization}.
Therefore, current~\enquote{ML as a Service}~(MLaaS) applications usually send and hence leak client data to a remote server operated by the service provider to perform inference~(cf.~top of~Figure~\ref{fig:ml_dummies}).

Even if the service provider and the remote server are considered trustworthy, the privacy of clients can still be compromised by breaches, hacks, and negligent or malicious insiders.
Such incidents occur frequently even at high-profile companies:
recently, Microsoft Outlook was hacked~\cite{outlook_2019} and AT\&T's customer support was bribed~\cite{att_2019}.
Thus, it is not enough to protect client data just from outsiders, it must also be hidden from the server to ensure privacy.

Previously, protecting the identity of individuals via anonymization techniques was seen as sufficient when learning on or inferring from data of a collection of users.
Such techniques reduce raw data to still enable extraction of knowledge without individuals being identifiable.
However, recent works conclude that current de-identification measures are insufficient and unlikely satisfy~GDPR standards~\cite{rocher_estimating_2019}.

\paragraph{Cryptography for ML Applications}
We believe this indicates that cryptographic measures
%of data protection 
should be employed to satisfy today's privacy demands.
The cryptographic literature has actively developed protocols and frameworks for efficient and privacy-preserving~ML in the past years. 
So far, efforts were focused on deep/convolutional neural networks, see~\cite{riazi_deep_2019} for a recent systematization of knowledge.
There, usually a scenario is considered where the server holds a model and performs~\emph{private}~ML inference on a client's data, with~\emph{no information} except for the inference result being revealed to the client~(cf.\ bottom of~Figure~\ref{fig:ml_dummies}).

Existing frameworks mostly rely on homomorphic encryption~(HE), secure multi-party computation~(SMPC), or a combination of both, to enable private inference with various security, resource, and usage properties.
As many~ML tasks today already require intense computational resources, the overhead incurred by introducing cryptographic privacy mechanisms is substantial.
Though a line of prominent frameworks from~CryptoNets~\cite{cryptonets} to~XONN~\cite{riazi_xonn:_2019} has established increased efficiency and relatively low execution times for private inference, research has~\emph{mainly focused on NNs} by looking for efficient ways to securely compute common activation functions, sometimes degrading accuracy by using more efficient approximations.
Existing frameworks only possess a low degree of automation and often require very low-level model descriptions, making it hard for non-experts to run private inference using their own models. 
Additionally, for approaches using SMPC, it is very common that the topology of the NN is leaked, which might reveal some model information to the client.

\paragraph{Our Contributions}
In this work, we present foundations and tools for privacy-preserving~ML in the unexplored domain of sum-product networks~(SPNs). Our framework, which we call~\emph{\coolname{}}, demonstrates that~SPNs can very well be protected with cryptographic measures.
Specifically, after presenting the necessary background for private ML and SPNs~(Section~\ref{sec_background}), we show how to efficiently perform private~SPN inference using SMPC~(Section~\ref{sec_privspn}).
We combine techniques from both~AI and applied cryptography to achieve this.
Contrary to popular SMPC-based approaches for protecting~NNs, ours leaks no information \CHANGED{from} the network topology by using Random Tensorized SPNs~(RAT-SPNs)~\cite{peharz2019random}.
We implement~\coolname{} using the state-of-the-art~SMPC framework~ABY~\cite{demmler2015} and provide an open-source tool that can transform~SPN instances from the~SPFlow framework~\cite{molina_spflow_2019} into privacy-preserving executables~(Section~\ref{sec_implementation}).
\coolname{} is easily usable by non-experts and intended to make private~ML available to the broader~AI community working on a wide range of sophisticated models.
In an experimental evaluation (Section~\ref{sec_eval}), we show that~\coolname{} performs private inference in reasonable time while preserving accuracy.
With our work, we push private~ML beyond~NNs and bring attention to the crucial, emerging task of making a~\emph{variety} of~ML applications private. 

\section{BACKGROUND}
\label{sec_background}
We start with the necessary background on secure computation, existing privacy-preserving ML solutions, and SPNs.

\subsection{Secure Computation (SC) of ML Tasks}

First described by~\cite{yao_generate_1986}, the concept of secure computation~(SC) lets computational parties~(e.g., a client and a server) evaluate arbitrary functions on secret inputs without leaking any information but the results.
For example, a server can calculate statistics on client data without learning the raw data, or a group of clients can jointly schedule meetings without revealing their availability. 
The~SC research community has put forth efficient schemes with practical implementations for applications that rely on homomorphic encryption~(HE) or secure multi-party computation~(SMPC).
The former allows computations directly on encrypted data, whereas in~SMPC an interactive protocol is executed between parties that, in the end, reveals only the desired output.
A general rule of thumb is that~SMPC requires more communication, whereas computation is the bottleneck for~HE.
In this work, we rely on secure \emph{two-party} computation, i.e., SMPC with two parties: client and server. 

\subsubsection{Privacy-Preserving Machine Learning}
We shortly recapitulate the most influential works for preserving privacy when performing machine learning tasks using SC techniques.

Privacy-preserving neural network inference was first proposed in~\cite{orlandi2007oblivious,sadeghi_generalized_2008,barni_privacy-preserving_2011}.
Secure classification via hyper-plane decision, naive~Bayes, and decision trees was presented in~\cite{goldwasser_ml}.
SecureML~\cite{mohassel_secureml:_2017} provides SMPC-friendly linear regression, logistic regression, and neural network training using~SGD as well as secure inference.
With CryptoNets~\cite{cryptonets}, the race for the fastest~NN-based privacy-preserving image classification began:
Mini\-ONN~\cite{liu_oblivious_2017}, Chameleon~\cite{riazi_chameleon:_2018}, Gazelle~\cite{juvekar_gazelle:_2018}, XONN~\cite{riazi_xonn:_2019}, \CHANGED{and DELPHI~\cite{delphi}} are only some of the proposed frameworks.

These frameworks mostly offer privacy-preserving deep/convolutional neural network inference based on~HE or~SMPC protocols, or even combinations of both techniques in different computational and security models.
However, they are not readily accessible to ML experts, as prototype implementations are not well-integrated into ML frameworks and require extensive cryptographic knowledge to secure applications.
Moreover, these frameworks are often engineered towards delivering outstanding performance for benchmarks with certain standard data sets~(e.g., MNIST~\cite{mnist}), but fail to generalize in terms of accuracy and performance.
There are some but very few attempts to directly integrate privacy technology into ML frameworks: for TensorFlow there exists rudimentary support for differential privacy~\cite{tfp}, HE~\cite{sealion}, and~SMPC~\cite{dahl_private_2018}, and for Intel's nGraph compiler there exists an HE backend~\cite{ngraph-he}.
Very recenty, Facebook's AI researchers released~CrypTen~\cite{crypten}, which provides an~SMPC integration with~PyTorch.
However, currently not much is known about the underlying cryptographic techniques and, therefore, its security guarantees.

Trusted execution environments~(TEEs) are an intriguing alternative to cryptographic protocols.
They use hardware features to shield sensitive data processing tasks.
TEEs are widely available, e.g., via~Intel Software Guard Extensions~(SGX), and therefore are explored for efficiently performing ML tasks~\cite{tensorscone}.
Unfortunately, Intel SGX provides no provable security guarantees and requires software developers to manually incorporate defenses against software side-channel attacks, which is extremely difficult.
Moreover, severe attacks on Intel~SGX allowed to extract private data from the TEE, making SGX less secure than cryptographic protocols~\cite{sgxpectre}.

\subsubsection{SMPC}
In SMPC, the function~$f$ to be computed securely is represented as a~Boolean circuit consisting of~XOR and~AND gates:
each gate is securely computed based on the encrypted outputs of preceding gates, and only the values of the output wires of the entire circuit are decrypted to obtain the overall output.
The intermediate results leak no information and only the outputs are decrypted by running a corresponding sub-protocol.

The literature considers two security settings:~\emph{semi-honest}, where the involved parties are assumed to honestly follow the protocol but want to learn additional information about other parties' inputs, and~\emph{malicious}, which even covers active deviations from the protocol.
SMPC protocols are usually divided into two phases: a~\emph{setup} phase that can be performed independently of the inputs (e.g., during off-peak hours or at night), and an \emph{online} phase that can only be executed once the inputs are known.
Most of the~\enquote{expensive} operations can be performed in the setup phase in advance such that the online phase is very efficient.
Two prominent~SMPC protocols are~Yao's garbled circuit~(GC)~\cite{yao_generate_1986} and the~GMW protocol~\cite{goldreich_play_1987}.
As we heavily rely on floating-point operations with high-depth circuits, we use Yao's GC protocol, which has a constant round complexity~(the round complexity of the~GMW protocol depends on the circuit depth and, hence, is not suited for our case).

\subsubsection{Yao's GC}
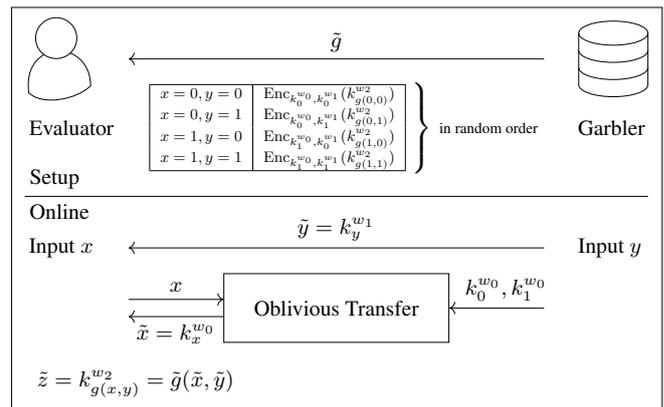
\begin{figure}[htb]
    \centering
    \resizebox{\columnwidth}{!}{%
    \begin{tikzpicture}[framed]
        \node [alice, minimum height=1.2cm, minimum width=0.5cm] (alice) at (-4, 0) {};

        \node [] (inpA) at (-3.825,-1) {Evaluator};

        \node [database, database radius=0.5cm, database segment height=0.25cm] (server) at (4,0) {};
        
        \node [] (inpB) at (3.975,-1) {Garbler};

        \draw [->] (3,0) -- (-3,0) node[above,pos=0.5] {$\tilde{g}$};

        \node [] (garbledT) at (-0.85,-1) {
            \scalebox{0.7}{
            \begin{tabular}{|l | l|}
                \hline
                $x = 0, y = 0$ & $\operatorname{Enc}_{k_0^{w_0}, k_0^{w_1}}(k_{g(0,0)}^{w_2})$ \\
                $x = 0, y = 1$ & $\operatorname{Enc}_{k_0^{w_0}, k_1^{w_1}}(k_{g(0,1)}^{w_2})$ \\
                $x = 1, y = 0$ & $\operatorname{Enc}_{k_1^{w_0}, k_0^{w_1}}(k_{g(1,0)}^{w_2})$ \\
                $x = 1, y = 1$ & $\operatorname{Enc}_{k_1^{w_0}, k_1^{w_1}}(k_{g(1,1)}^{w_2})$ \\
                \hline
            \end{tabular}
            }
        };
        
        \node (parenR) at (1.25,-1) {$\left\}\vphantom{\rule{0cm}{0.75cm}}\right.$};
        \node (parenRLabel) at (2.2,-1) {\scalebox{0.7}{in random order}};

        \draw [] (4.5,-2) -- (-4.5,-2) node[above,xshift=0.5cm] {Setup\phantom{e}} node[below,xshift=0.5cm] {Online};
        
        \node [] (inpA) at (-3.975,-2.75) {Input $x$};
        \node [] (inpB) at (3.95,-2.75) {Input $y$};
        
        \draw [->] (3,-2.75) -- (-3,-2.75) node[above,pos=0.5] {$\tilde{y} = k_y^{w_1}$};
        
        \node [shape=rectangle,draw,minimum width=3.25cm, minimum height=1cm] (OT) at (0, -3.625) {Oblivious Transfer};
        
       \draw [->] (3,-3.625) -- (OT.east) node[above,pos=0.4] {$k_0^{w_0}, k_1^{w_0}$};
       \draw [->] (-3,-3.5) -- ([yshift=0.125cm]OT.west) node[above,pos=0.5] {$x$};
       \draw [->] ([yshift=-0.125cm]OT.west) -- (-3,-3.75) node[below,pos=0.5] {$\tilde{x}=k_x^{w_0}$};
        
       \node [] (comp) at (-2.9,-4.725) {$\tilde{z} = k_{g(x, y)}^{w_2} = \tilde{g}(\tilde{x}, \tilde{y})$};
    \end{tikzpicture}
    }
    \vspace{-2em}
    \caption{Overview of Yao's GC protocol for securely evaluating a binary gate~$g$ (output wire~$w_2$) on binary inputs~$x$~(input wire~$w_0$) and~$y$ (input wire~$w_1$), where~$\tilde{g}$ is the garbled truth table. All labels~$k_a^w$~(corresponding to wire~$w$ having value~$a$) are generated uniformly at random by the garbler. The protocol is composable and can be used to securely compute any circuit based on the secure computation of one gate.} 
    \label{fig:yao}
\end{figure}

We present a schematic overview for the secure evaluation of a single gate in Figure~\ref{fig:yao} and refer to~\cite{lindell2009proof} for further technical details.

The central idea of this protocol is to~\emph{encode} the function~(more precisely, its representation as a~Boolean circuit) and the inputs such that the encoding reveals no information about the inputs but can still be used to evaluate the circuit.
This encoding is called~\enquote{garbling} and individual garbled values can be seen as encryptions.
We will use the common notation of~$\tilde{g}$ and~$\tilde{x},\tilde{y}$ to refer to a garbled gate or garbled inputs, respectively.
The evaluation of the garbled circuit~$\tilde{C}$~(consisting of many garbled gates) using the garbled inputs, in turn, results in an encoding~$\tilde{z}=\tilde{C}(\tilde{x},\tilde{y})$ of the output.
The encoded output can only be decoded~\emph{jointly} to the plain value~$z=C(x,y)$, i.e., both parties have to agree to do so.
In the protocol, one of the parties, the~\enquote{garbler}, is in charge of creating the garbled circuit.
The other party, the~\enquote{evaluator}, obtains the garbled circuit, evaluates it, and then both parties jointly reveal the output.

\paragraph{Circuit Garbling}
The wires in the Boolean circuit~$C$ of a function~$f$ are assigned two~\emph{randomly} chosen labels~/ keys:~$k_0$ and~$k_1$, indicating that the wire's plain value is~$0$ or~$1$, respectively.
Though there is a label for each possible value~$\{0,1\}$, only the garbled value~$\tilde{x}=k^w_x$ for the actual plain value~$x$ of wire~$w$ is used to evaluate the garbled circuit.
The garbler creates both labels and therefore is the only one who knows the mapping to plaintext values~-- for the evaluator, a single randomly-looking label reveals no information.

The garbler creates a randomly permuted~\enquote{garbled} gate~$\tilde{g}$ in the form of an encrypted truth table for each gate~$g$ in the circuit~$C$ of~$f$, and sends all garbled gates to the evaluator.
The key idea is to use an encryption scheme~$\operatorname{Enc}$ that has~\emph{two} encryption keys.
For each truth table entry, the label associated with the plaintext value of the outgoing wire is then encrypted using the labels associated with the plain values of the two incoming wires as encryption keys~(cf.~Figure~\ref{fig:yao}). %labels associated with the respective incoming wire labels as encryption keys.

\paragraph{Garbled Circuit Evaluation}
Now, if the evaluator is in possession of the labels~$\tilde{x}$ and~$\tilde{y}$ corresponding to the incoming wires' values~$x$ and~$y$, then exactly one entry of~$\tilde{g}$ can be successfully decrypted using~$\tilde{x}$ and~$\tilde{y}$ as decryption keys\footnote{A special type of encryption scheme is used to detect if the decryption was successful or not.}.
This will result in~$\tilde{z} = \tilde{g}(\tilde{x},\tilde{y})$, the label of the outgoing wire of~$g$ associated with the desired plaintext value~$z = g(x,y)$.
Since only the desired entry can be decrypted and, given that the labels are chosen randomly and independently of the wire values, the evaluator can perform this computation without learning any plaintext information.

The remaining challenge is that the evaluator needs to obtain the correct garbled inputs~(i.e., the labels corresponding to its inputs) without revealing the inputs to the garbler.
This is solved by a cryptographic protocol called oblivious transfer~(OT), which enables one party with input bit~$b$ to obliviously obtain a string~$s_b$ from another party holding two input strings~$s_0,s_1$ without revealing~$b$ and learning anything about~$s_{1-b}$.
With this building block, Yao's GC protocol is composed as follows~(cf.~Figure~\ref{fig:yao}):
In the setup phase, the garbler creates all wire labels for the garbled circuit~$\tilde{C}$ and sends~$\tilde{C}$ to the evaluator.
During the online phase, the garbler sends the labels corresponding to its input to the evaluator.
The evaluator's garbled inputs are obtained via~OT.
Then, the evaluator decrypts~$\tilde{C}$.
The output can be jointly decrypted if the parties reveal the output label associations.

\paragraph{Protocol Costs}
Improvements on OT~\cite{ishai_extending_2003, asharov_more_2017} and the garbling scheme~\cite{kolesnikov_improved_2008,zahur_two_2015} have significantly reduced the overhead of~Yao's~GC protocol, making it viable to be used in applications. 
Specifically, the GC created and sent in the setup phase requires~$2\kappa$ bits per binary AND gate, where~$\kappa$ is the symmetric security parameter~(e.g., $\kappa=128$ for the currently recommended security level).
For obliviously transferring the labels corresponding to the evaluator's input~$x$, $|x|\kappa$ bits must be sent in the setup phase, as well as~$|x|(2\kappa + 1)$ bits in the online phase.
Additionally, the labels for the garbler's input~$y$ must be sent in the online phase~($|y|\kappa$ bits).
The protocol only requires a constant number of rounds of interaction.

\subsection{Sum-Product Networks (SPNs)}

Recent years have seen a significant interest in tractable probabilistic representations such as Arithmetic Circuits~(ACs)~\cite{choiD17}, Cutset Networks~\cite{rahman2014cutset}, and~SPNs~\cite{Poon2011}. 
In particular, SPNs, an instance of~ACs, are deep probabilistic models that can represent high-treewidth models~\cite{Zhao2015} and facilitate~\emph{exact} inference for a range of queries in time~\emph{linear} in the network size.

\subsubsection{Definition of SPNs} 
Formally, an SPN is a rooted directed acyclic graph, consisting of~\emph{sum}, \emph{product}, and~\emph{leaf} nodes.
The scope of an SPN is the set of random variables~(RVs) appearing on the network.
An~SPN can be defined recursively as follows:~(1)~a tractable univariate distribution is an~SPN;~(2)~a product of SPNs defined over different scopes is an SPN; and~(3)~a convex combination of SPNs over the same scope is an SPN. 
Thus, a product node in an SPN represents a factorization over independent distributions defined over different~RVs~$P(\bm{X},\bm{Y}) = P(\bm{X})P(\bm{Y})$, while a sum node stands for a mixture of distributions defined over the same variables~$P(\bm{X},\bm{Y}) = w P_1(\bm{X},\bm{Y}) + (1-w) P_2(\bm{X},\bm{Y})$.
From this definition, it follows that the joint distribution modeled by such an~SPN is a valid normalized probability distribution~\cite{Poon2011}.

\subsubsection{Tractable Inference in SPNs}
To answer probabilistic queries in an SPN, we evaluate the nodes starting at the leaves. 
Given some evidence, the probability output of querying leaf distributions is propagated bottom up.
For product nodes, the values of the child nodes are multiplied and propagated to their parents.
For sum nodes, we sum the weighted values of the child nodes.
The value at the root indicates the probability of the asked query.
To compute marginals, i.e., the probability of partial configurations, we set the probability at the leaves for those variables to~$1$ and then proceed as before.
This allows us to also compute conditional queries such as~$P(\bm{Y}|\bm{X}) = P(\bm{X},\bm{Y}) / P(\bm{X})$.
Finally, \CHANGED{using a bottom-up and }top-down pass, \CHANGED{we can compute approximate~MPE states~\cite{Poon2011}}.
All these operations traverse the tree at most twice and therefore can be achieved in linear time~w.r.t.\ the size of the SPN.

\section{\coolname{} INFERENCE}
\label{sec_privspn}
Given the AC structure of~SPNs, SMPC is a fitting mechanism to preserve privacy in SPN inference as it relies on securely evaluating a circuit.
\CHANGED{Compared to,~e.g.,~NNs,~SPNs do not have alternating linear and non-linear layers, which would complicate the application of~SMPC protocols.}
Here, we are concerned with \emph{private}~SPN inference in a setting where the client has a private input and the server is in possession of a model; in the end, the server learns nothing, and the client only learns the inference result~(cf.~bottom of Figure~\ref{fig:ml_dummies}).
Unfortunately, we cannot use the arithmetic version of the~GMW protocol, as it only provides integer or fixed-point operations, which is insufficient for tractable and normalized probabilistic inference such as the case of SPNs. 
Instead, \coolname{} uses Yao's GC protocol that evaluates Boolean circuits, which allows us to use floating point operations by including Boolean sub-circuits corresponding to IEEE~754-compliant~$32$- or~$64$-bit floating point operations~\cite{demmler_automated_2015} in the circuit representation of the to-be-evaluated SPN.

\subsection{Secure Evaluation of SPNs with SMPC}

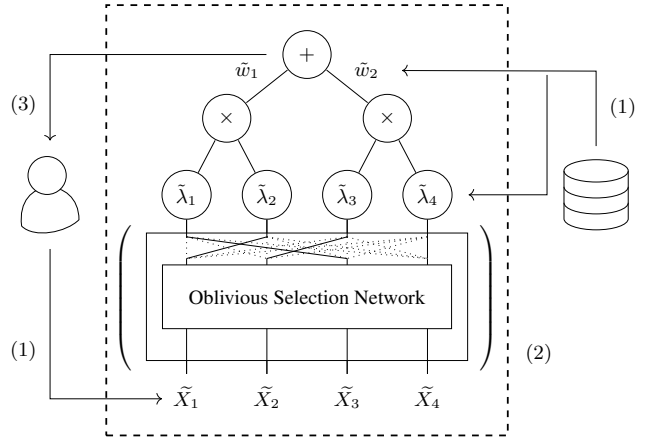
\begin{figure}[t]
    \centering
    \resizebox{\columnwidth}{!}{%
    \begin{tikzpicture}[yscale = 0.8]
        \tikzstyle{circ} = [draw, circle, minimum size=0.75cm]
        \tikzstyle{inp} = [minimum size=0.75cm]
        \tikzstyle{conn} = [shorten >=0.39cm,shorten <=0.39cm]
        \tikzstyle{conn2} = [shorten >=0.39cm,shorten <=2.1cm]
        \tikzstyle{conn3} = [shorten >=2.53cm,shorten <=0.39cm]
        \tikzstyle{conn4} = [shorten >=2.2cm,shorten <=1.1cm]
        \tikzstyle{osn} = [dotted]
    
        \node [circ] (sum) at (0,0) {$+$};
        
        \node [circ] (prod1) at (-1.25,-1.25) {$\times$};
        \node [circ] (prod2) at (1.25,-1.25) {$\times$};
        
        \draw [conn] (sum.center) -- (prod1.center) node[left,pos=0.5,yshift=0.25cm] {$\tilde{w}_1$};
        \draw [conn] (sum.center) -- (prod2.center) node[right,pos=0.5,yshift=0.25cm] (w1) {$\tilde{w}_2$};
        
        \node [circ] (dist1) at (-1.875,-2.75) {$\tilde{\lambda}_1$};
        \node [circ] (dist2) at (-0.625,-2.75) {$\tilde{\lambda}_2$};
        \node [circ] (dist3) at (0.625,-2.75) {$\tilde{\lambda}_3$};
        \node [circ] (dist4) at (1.875,-2.75) {$\tilde{\lambda}_4$};
        
        \draw [conn] (prod1.center) --  (dist1.center);
        \draw [conn] (prod1.center) -- (dist2.center);
        
        \draw [conn] (prod2.center) -- (dist3.center);
        \draw [conn] (prod2.center) -- (dist4.center);
        
        \node [inp] (inp1) at (-1.875,-6.75) {$\fixwidetilde{X_1}$};
        \node [inp] (inp2) at (-0.625,-6.75) {$\fixwidetilde{X_2}$};
        \node [inp] (inp3) at (0.625,-6.75) {$\fixwidetilde{X_3}$};
        \node [inp] (inp4) at (1.875,-6.75) {$\fixwidetilde{X_4}$};
        
        \draw [conn] (dist1.center) -- (inp1.center);
        \draw [conn] (dist2.center) -- (inp2.center);
        \draw [conn] (dist3.center) -- (inp3.center);
        \draw [conn] (dist4.center) -- (inp4.center);
        
        \node [draw, rectangle, fill=white, minimum height=2cm, minimum width=5cm] (sel) at (0,-4.75) {Oblivious Selection Network};
        
        \node [draw, rectangle, fill=white, minimum height=1cm, minimum width=4.5cm] (selin) at (0,-4.75) {Oblivious Selection Network};
        
        \draw [conn2] (dist1.center) -- (inp1.center);
        \draw [conn2] (dist2.center) -- (inp2.center);
        \draw [conn2] (dist3.center) -- (inp3.center);
        \draw [conn2] (dist4.center) -- (inp4.center);
        
        \draw [conn3] (dist1.center) -- (inp1.center);
        \draw [conn3] (dist2.center) -- (inp2.center);
        \draw [conn3] (dist3.center) -- (inp3.center);
        \draw [conn3] (dist4.center) -- (inp4.center);
        
        \draw [conn4] (dist1.center) -- (inp1.center);
        \draw [conn4] (dist2.center) -- (inp2.center);
        \draw [conn4] (dist3.center) -- (inp3.center);
        \draw [conn4] (dist4.center) -- (inp4.center);
        
        \draw [osn] (-1.875,-3.575) -- (-1.875,-4);
        \draw [] (-0.625,-3.575) -- (-1.875,-4);
        \draw [osn] (0.625,-3.575) -- (-1.875,-4);
        \draw [osn] (1.875,-3.575) -- (-1.875,-4);
        
        \draw [osn] (-1.875,-3.575) -- (-0.625,-4);
        \draw [osn] (-0.625,-3.575) -- (-0.625,-4);
        \draw [] (0.625,-3.575) -- (-0.625,-4);
        \draw [osn] (1.875,-3.575) -- (-0.625,-4);
        
        \draw [osn] (-1.875,-3.575) -- (1.875,-4);
        \draw [osn] (-0.625,-3.575) -- (1.875,-4);
        \draw [osn] (0.625,-3.575) -- (1.875,-4);
        \draw [] (1.875,-3.575) -- (1.875,-4);
        
        \draw [] (-1.875,-3.575) -- (0.625,-4);
        \draw [osn] (-0.625,-3.575) -- (0.625,-4);
        \draw [osn] (0.625,-3.575) -- (0.625,-4);
        \draw [osn] (1.875,-3.575) -- (0.625,-4);
        
        \node (parenL) at (-2.85,-4.75) {$\left(\vphantom{\rule{0cm}{1.25cm}}\right.$};
        \node (parenR) at (2.85,-4.75) {$\left)\vphantom{\rule{0cm}{1.25cm}}\right.$};
        
        \node [draw, thick, dashed, rectangle, minimum height=6.75cm, minimum width=6.25cm, label={[xshift=0.15cm,yshift=-2.1cm]right:$(2)$}] (box) at (0,-3.25) {};
        
        \node [database, database radius=0.5cm, database segment height=0.25cm] (server) at (4.5,-2.75) {};
        
        \node [alice, minimum height=1.2cm, minimum width=0.5cm] (alice) at (-4, -2.75) {};
        
        \draw [->, shorten <=0.25cm] (alice.south) |- (inp1.west) node[circle, left, pos=0.5, yshift=0.75cm] {$(1)$};

        \draw [->, shorten >=0.25cm] ([shift={(-0.25cm,0cm)}]sum.west) |- ([shift={(-4cm,0cm)}]sum.center) -- (alice.north) node[circle, left, pos=0.5, yshift=0.125] {$(3)$};
        
        \draw [->] ([shift={(-0.25cm,2.35cm)}]server.west) -- ([shift={(-0.25cm,0cm)}]server.west) |- ([shift={(0.25cm,0cm)}]dist4.east);

        \draw [->, shorten >=0.25cm] ([shift={(0cm,0.25)}]server.north) |- ([shift={(0cm,2.425)}]server.center) node[circle, right, pos=0.25, yshift=0.125] {$(1)$} -- (w1.east) ;
    \end{tikzpicture}
    }
    \vspace{-2em}
    \caption{\coolname{} protocol flow for an exemplary miniature SPN with~Poisson leaves: (1)~client and server have private inputs~$X_{1,\dots,4}$ and~$w_{1,2},\lambda_{1,\ldots,4}$, respectively;~(2)~private evaluation of leaf-, sum- and product nodes using~SMPC;~(3)~client receives~SPN inference result.}
    \label{fig:spn_stpc}
\end{figure}

Our approach~(cf. Figure~\ref{fig:spn_stpc}) is to transform the SPN into a Boolean circuit and to then evaluate it via SMPC.
The server input consists of all the model parameters of the SPN~(i.e., weights for the weighted sums and parameters for the leaf distribution), the client input consists of the evidence, and the output is the root node value.
We perform all computations in the log-domain using the well-known~\emph{log-sum-exp} trick, which also provides a runtime advantage for our~SMPC approach as it replaces products with more efficient additions.
Contrary to the convention, we use the~log2 domain in~\coolname{} since the circuits for log2 and exp2 operations are significantly smaller than the natural~log and~exp operations.

Due to the~SMPC security properties, all the model parameters are hidden from the client and the input values or evidence from the server.
However, this naive approach alone does not provide our desired privacy guarantees since the circuit evaluated in~SMPC is public.
Therefore, the topology of the~SPN is leaked to the client, including which~RVs~(the scope) are used in which leaves.
Depending on how the SPN was learned, this might reveal information about the server's model, such as correlations among~RVs, number of mixtures, etc. %to the client
To hide this information, one could make use of generic~\emph{private function evaluation} techniques such as incorporating~\emph{universal circuits}~(UCs)~\cite{valiant_universal_1976}.
UCs allow one party to choose a function as the private input, which is then obliviously evaluated on the other party's input such that nothing about the function or the input is revealed.
Employing these generic techniques, however, would drastically increase the overhead we introduce via~SMPC.
For this reason, the related work on~SMPC for private~NN inference usually assumes that the~NN topology is public, with the impact on model privacy being unclear in this situation.
To mitigate these concerns in~\coolname{}, we tailor efficient techniques stemming from both~AI as well as applied cryptography research specifically to~SPNs.
The first method hides specifics of the training data by using Random Tensorized SPNs~(RAT-SPNs)~\cite{peharz2019random}, while the second method allows to hide the scope of any existing~SPN without the need to re-learn a~RAT-SPN.

\subsubsection{Hiding the Training Data}

It is possible that the structure of a general SPN leaks information about the training data. To hide any information that could be revealed from the~SPN structure, we propose to use RAT-SPNs~\cite{peharz2019random}. 
The RAT-SPN structure is built randomly via region graphs.
Given a set of~RVs~$\bm{X}$, a \emph{region}~$\region$ is defined as any non-empty subset of~$\bm{X}$.
Given any region~$\region$, a~$K$-\emph{partition}~$\partition$ of~$\region$ is a collection of~$K$ non-overlapping sub-regions~$\region_1, \dots, \region_K$, whose union is~$\region$, i.e.,~$\partition = \{\region_1, \dots, \region_K\}$, $\forall k \colon \region_k \not= \emptyset$, $\forall k \not= l \colon \region_k \cap \region_l = \emptyset$, $\bigcup_k \region_k = \region$.
This partitioning algorithm randomly splits the RVs.
Furthermore, we recursively split the regions until we reach a desired partitioning depth.
Here, we consider only 2-partitions.
From these region graphs, we can construct an SPN specifying the number of uniform leaves per~RV.
Since the structure-building algorithm is data-agnostic (it only knows the number of~RVs in the dataset), there is no information leakage. This also means that any initial random structure for~$|\bm{X}|$, the number of random variables, is a valid initial structure for any other dataset with the same number of dimensions.
After obtaining the structure, we use a standard optimization algorithm for parameter estimation. 
The structure produced by the RAT-SPN algorithm is regular, and the values of the parameters after the optimization encode the knowledge needed to build the joint distribution. In our scheme, the parameters are only visible to the service provider.
Using a random structure also enables us to choose the size of the~SPNs, which allows service providers to trade off model complexity, efficiency, and accuracy.

\subsubsection{Hiding the Scope}
\label{subsubsec:privspn_scope}

Since the scope of a node is defined by the scope of its children, it suffices to hide the leaves' scopes. 
Concretely, for each leaf, we have to hide which~$X_j$ for~$j\in\{1, \ldots, n\}$ from the client's~RVs~$\bm{X} = X_1, \ldots, X_n$ is selected.
This corresponds to an~\emph{oblivious array access} in each leaf, where an array can be accessed without revealing the accessed location~$j$. 
There exist efficient methods to do this based on homomorphic encryption~\cite{brickell_privacy_2007} or~secure evaluation of selection networks~\cite{kolesnikov_practical_2008} via SMPC.
A recent study of private decision tree evaluation~\cite{kiss_sok:_2019} shows that selection networks outperform selection based on~HE in both total and online runtime.
Hence, we obliviously select~RVs via securely evaluating a selection network in~\coolname{}.

Similar to the usage in decision trees, we add just one selection network below the~SPN instead of selecting one variable per leaf.
That is, the variable input of the secure leaf computation~(see below) is the outcome of the selection network, which selects the variables~$X_{\phi(1)}, \ldots, X_{\phi(m)}$ for the~$m$ leaves in the~SPN from the~$n$ client inputs according to a server input~$\phi\colon [m] \mapsto [n]$ denoting which leaf~$i\leq m$ uses~$X_{\phi(i)}$.
If~$m \geq n$~(which we assume is true since~RVs are usually used more than once), the complexity of such a selection network is~\cite{kolesnikov_practical_2008}:
\begin{equation*}
    C^{\text{sel}}_{n,m} = \frac{1}{2} \left(n + m\right)\log_2(n) + m\log_2(m) - n + 1,
\end{equation*}
beating the trivial solution of $O(nm)$.
This requires~$|X|\kappa (n + 2C^{\text{sel}}_{n,m})$ bits of setup communication and~$|X|n(2\kappa + 1)$ bits online~\cite{kiss_sok:_2019}.
Hereby, one can hide the scope of any~SPN, including ones learned through other methods~\cite{gens2013learning}~(although the topology is still leaked).
We propose to use this approach to increase privacy in cases where leaking the topology is deemed acceptable, or where re-learning the structure of an already existing~SPN is infeasible. % deemed

\subsubsection{Hiding the RVs and Leaf Parameters}
Because the secure computation of each floating point operation introduces overhead, our approach at the leaf level is to let the respective parties locally pre-compute as many terms as possible before inputting them into the secure SPN evaluation.
For Gaussians in the~log2 domain, the result can be evaluated in SMPC with just two multiplications and two additions based on the client's~RV input~$X_j$ and server inputs~$\mu$,~$s_1 = -\log_2(2\pi \sigma^2)$, and~$s_2 = \frac{\log_2(e)}{2\sigma^2}$ based on parameters mean~$\mu$ and variance~$\sigma^2$:
\begin{equation*}
    \fixwidetilde{P}_{\text{Gauss}}\left(X_j; \mu, \sigma^2\right) = \tilde{s}_1 - \left( \fixwidetilde{X_j}-\tilde{\mu}\right)^2 \tilde{s}_2.
\end{equation*}
Thus, for each leaf, the~SPN circuit requires~$C^{\text{Gauss}}_b = 2(C^{\text{ADD}}_b + C^{\text{MUL}}_b)$~AND gates, where~$C^{\text{OP}}_b$ denotes the number of~AND gates for a~$b$-bit floating point operation~OP, cf.~\cite{demmler_automated_2015}\footnote{For instance, $C^{\text{ADD}}_{32}=\si{1820}, C^{\text{MUL}}_{32}=\si{3016}, C^{\text{EXP2}}_{32}=\si{9740}$, and $C^{\text{LOG2}}_{32}=\si{10568}$.}.
Additionally, for the entire~SPN,~$\textit{IC\kern 1pt}^{\text{Gauss}}_b = nb$ bits of client input and~$\textit{IS\kern 1pt}^{\text{Gauss}}_b = 3mb$ bits of server input are added, where~$m$ is the amount of leaves and~$n$ is the number of~RVs. 

Similarly, we can securely compute Poissons with just one multiplication and two additions based on the client's~RV inputs~$X_j$ and~$c_1 = -\log_2(X_j!)$, and server inputs~$s_1 = \log_2(\lambda)$ and~$s_2 = -\lambda\log_2(e)$ based on mean~$\lambda$:
\begin{equation*}
    \fixwidetilde{P}_{\text{Pois}}\left(X_j; \lambda\right) = \fixwidetilde{X_j} \cdot \tilde{s}_1 + \tilde{c}_1 + \tilde{s}_2.
\end{equation*}
This results in the leaf size~$C^{\text{Pois}}_b = 2C^{\text{ADD}}_b + C^{\text{MUL}}_b$ with input sizes~$\textit{IC\kern 1pt}^{\text{Pois}}_b = 2nb$ and~$\textit{IS\kern 1pt}^{\text{Pois}}_b = 2mb$ for the entire SPN. 

\iffull Bernoullis consist of just one MUX gate, selecting from two server inputs~$p$ and~$1-p$ based on the binary client variable input. This requires $|p|$ AND gates or, when using the GMW protocol, MUX gates can also be optimized via OTs and require only~$2\kappa$ bits of communication in the setup phase and $4|p|$ bits in the online phase.
\else 
Bernoullis consist of just one~MUX gate, selecting from two server inputs~$p$ and~$q = 1-p$ based on the binary client RV input~$X_j$:
\begin{equation*}
    \fixwidetilde{P}_{\text{Bern}}\left(X_j; p\right) = \text{MUX}\left(\fixwidetilde{X_j}, \tilde{p}, \tilde{q}\right) = \begin{cases}
      \tilde{p}, & \text{if}\ X_j=1 \\
      \tilde{q}, & \text{if}\ X_j=0
    \end{cases}.
\end{equation*}
Hence, they have a complexity of~$|p|$ AND gates, yielding the costs~$C^{\text{Bern}}_b = b$,~$\textit{IC\kern 1pt}^{\text{Bern}}_b = n$, and~$\textit{IS\kern 1pt}^{\text{Bern}}_b = 2mb$.

Due to the~log2 domain, computations of a product node just introduce a complexity of~$(ch(s)-1)C^{\text{ADD}}_b$, where~$ch(s)$ denotes the amount of children of a node~$s$.
For the same reason, the complexity of a sum node is:
\begin{equation*}
    C^{\text{LOG2}}_b + (ch(s)-1)C^{\text{ADD}}_b + ch(s)(C^{\text{ADD}}_b + C^{\text{EXP2}}_b).
\end{equation*}

\subsection{Efficiency}
\label{subsec:framework_efficiency}
Putting all of the presented building blocks together, we get the following amount of~AND gates~(the only relevant cost metric for~Yao's~GC protocol) for an~SPN with~$n$~RVs and~$m$ leaves of distribution~$D\in\{\text{Gauss}, \text{Pois}, \text{Bern}\}$ that operates with~$b$-bit precision and consists of a set of sum nodes~$\bm{S}$ and product nodes~$\bm{P}$, where~$ch(s)$ for~$s\in\bm{S}\cup\bm{P}$ denotes the amount of children of node~$s$:

\scalebox{0.9}{\parbox{.5\linewidth}{%
\begin{align*}
    C^{\text{SPN}}_b =& mC^D_b + \sum_{s\in\bm{S}} \left(C^{\text{LOG2}}_b + (ch(s)-1)C^{\text{ADD}}_b\right.\\
                 +&  \left. ch(s)\left(C^{\text{ADD}}_b + C^{\text{EXP2}}_b\right)\right)   + \sum_{p\in\bm{P}}\left(ch(p)-1\right)C^{\text{ADD}}_b.
\end{align*}
}}

In addition, we also have~$\textit{IC\kern 1pt}^D_b$ client input bits stemming from the RVs and~$\textit{IS\kern 1pt}^D_b + b\sum_{s\in\bm{S}}ch(s)$ server input bits stemming from the leaf parameters as well as the sum weights. Therefore, using~Yao's~GC protocol, \coolname{} has the following communication costs in bits in the~\emph{setup phase}:
\begin{equation*}
    \kappa \left( \textit{IC\kern 1pt}^D_b + 2C^{\text{SPN}}_b \right)
\end{equation*}
and in the~\emph{online phase}:
\begin{equation*}
    \kappa \left( 2\textit{IC\kern 1pt}^D_b + \textit{IS\kern 1pt}^D_b + b\sum_{s\in\bm{S}}ch(s)\right) + \textit{IC\kern 1pt}^D_b,
\end{equation*}
where~$\kappa$ is the symmetric security parameter~(e.g.,~$\kappa=128$).

If one does not use~RAT-SPNs and instead our scope-hiding private~SPN evaluation, obliviously selecting the leaves'~RVs for~Gaussian, Poisson, and~Bernoulli leaves has the following online communication, respectively:~$nb(2\kappa+1)$,~$2nb(2\kappa+1)$, and~$n(\kappa + 1)$ bits.
The setup communication is~$\kappa b\left(n + 2C^\text{sel}_{n,m}\right)$,~$\kappa b\left(2n + 2C^\text{sel}_{2n,m}\right)$, and~$\kappa\left(n + 2C^\text{sel}_{n,m}\right)$ bits, respectively.

\subsection{Security Guarantees}

\CHANGED{As we use~RAT-SPNs, the underlying structure, the size and depth of the~SPN are known to the client. However, this structure is randomly generated and comes only from hyper-parameters. Therefore, the structure is independent of training data and leaks no private information.
The number of random variables is known by both parties, as usual (e.g., \cite{cryptonets,juvekar_gazelle:_2018,liu_oblivious_2017,delphi,riazi_xonn:_2019,riazi_chameleon:_2018}).
And while the value of the input variables is hidden, the output is not and might reveal some information but is data that~\emph{inherently} has to be revealed.
}

The protocols we use in our implementation of \coolname{} are provably secure in the semi-honest model~\cite{lindell2009proof}.
In the studied setting, it is reasonable to assume the server is semi-honest, as reputable service providers are confined by regulations and potential audits.
Furthermore, detected malicious behaviour would hurt their reputation, providing an economic incentive to behave honestly.
However, these regulations and incentives do not exist for the client's device, which can be arbitrarily modified by the client or harmful software.

\CHANGED{Fortunately, CryptoSPN can easily be extended to provide security against malicious clients as it relies on~Yao's~GC protocol.
There, the only messages sent by the~(potentially malicious) client are in the oblivious transfer. Thus, one just needs to instantiate a maliciously secure~OT protocol to achieve security against malicious clients, which incurs only a negligible performance overhead~\cite{asharov_more_2017}.}

\section{IMPLEMENTATION \& SPFlow INTEGRATION}
\label{sec_implementation}

We implemented \coolname{} using the state-of-the-art~SMPC framework ABY~\cite{demmler2015} with the floating point operation sub-circuits of~\cite{demmler_automated_2015}
and the selection network circuit of~\cite{kiss_sok:_2019}.
ABY implements various~SMPC protocols in~C++ and provides~APIs for the secure evaluation of supplied circuits within these protocols.
It also supports~\emph{single instruction, multiple data}~(SIMD) instructions, which allows~\coolname{} to batch-process multiple queries at the same time.
Notably, like most other~SMPC frameworks, ABY requires a very low-level circuit description of the function that is computed securely, making it hard for~AI researchers and others without a background in cryptography to actually perform private~ML inference.
Motivated by this gap, we integrate \coolname{} with~SPFlow~\cite{molina_spflow_2019}, an open-source~Python library that provides an interface for~SPN learning, manipulation, and inference.
For users, \coolname{} appears as another~SPFlow export that enables private~SPN inference.
Specifically, \coolname{} allows~ML experts to easily transform an~SPN in SPFlow into a privacy-preserving~ABY program with just the~SPN as input.
The resulting~ABY program \CHANGED{can be }compiled into an executable
for simple deployment on the client and server side.
\CHANGED{\coolname{} is available at~\url{https://encrypto.de/code/CryptoSPN}.}

\section{EXPERIMENTAL EVALUATION}
\label{sec_eval}
We evaluate \coolname{} on random~SPNs trained with~SPFlow for the standard datasets provided in~\cite{gens2013learning}, and on regular~SPNs for~\texttt{nips}, a count dataset from~\cite{molina2017poisson}.
We evaluate models with both~$32$- and~$64$-bit precision to study the trade-off between accuracy and efficiency.
The experiments are performed on two machines with Intel Core~i9-7960X~CPUs and~\SI{128}{\giga\byte} of~RAM.
We use a symmetric security parameter of~$\kappa=128$ bits according to current recommendations.
The connection between both machines is restricted to~\SI{100}{\mega\bit\per\second} bandwidth and a round-trip time of~\SI{100}{\milli\second} to simulate a realistic wide-area network~(WAN) for a client-server setting.

\setlength{\aboverulesep}{0pt}
\setlength{\belowrulesep}{0pt}
\begin{table*}[ht]
\sisetup{round-mode=places, table-number-alignment=center, table-text-alignment=center}
\centering
\caption{Benchmarks of private SPN inference with \coolname{} in a~WAN network. The SPN has~$|\bm{X}|$~RVs,~$|\bm{S}|$ sum nodes, and~$|\bm{P}|$ product nodes. Setup and online runtime as well as communication are measured for both~$32$- and $64$-bit precision. All SPNs are~RAT-SPNs with~Bernoulli leaves except the ones for \texttt{nips}, which are regular~SPNs with~Poisson leaves. $\dagger$ indicates usage of a selection network for hiding RV assignments.}\smallskip
\vspace{-0.5em}
%\resizebox{0.95\textwidth}{!}{ % If your table exceeds the column or page width, use this command to reduce it slightly
\nprounddigits{0}
\begin{tabular}{l|S[table-format=4]S[table-format=2]S[table-format=4]S[table-format=5]S[table-format=5]S[table-format=2]|S[round-precision=0,table-format=3.0]S[round-precision=0,table-format=3.0]S[round-precision=2,table-format=1.2]S[round-precision=2,table-format=1.2]|S[round-precision=1,table-format=2.1]S[round-precision=1,table-format=2.1]S[round-precision=1,table-format=2.1]S[round-precision=1,table-format=2.1]}
\toprule
{\multirow{2}{*}{dataset}} & {\multirow{2}{*}{$|\bm{X}|$}} & {\multirow{2}{*}{$|\bm{S}|$}} & {\multirow{2}{*}{$|\bm{P}|$}} & {\multirow{2}{*}{\#leaves}} & {\multirow{2}{*}{\#edges}} & {\multirow{2}{*}{\#layers}} & \multicolumn{2}{c}{setup (s)} & \multicolumn{2}{c}{setup (GB)} & \multicolumn{2}{c}{online (s)} & \multicolumn{2}{c}{online (MB)} \\
\cmidrule(lr){8-9}\cmidrule(lr){10-11}\cmidrule(lr){12-13}\cmidrule(lr){14-15}
&&&&&&& {$32$b} & {$64$b} & {$32$b} & {$64$b} & {$32$b} & {$64$b} & {$32$b} & {$64$b} \\  
\midrule
\texttt{accidents} & 111 & 22 & 4420 & 11100 & 27161 & 7 & 364.790 & 824.759 & 4.344846884 & 9.827901661 & 22.483 & 55.546 & 15.476771 & 30.949928  \\ 
\texttt{baudio} & 100 & 22 & 4420 & 10000 & 26061 & 7 & 359.144 & 811.645 & 4.278353718 & 9.672070918 & 22.116 & 53.575 & 14.350017 & 28.696772  \\
\texttt{bbc} & 1058 & 2 & 880 & 42320 & 44721 & 5 & 247.946 & 577.277 & 2.945730031 & 6.868429682 & 18.644 & 41.921 & 43.779831 & 87.525628\\
\texttt{bnetflix} & 100 & 2 & 4400 & 20000 & 32001 & 5 & 264.369 & 604.219 & 3.145345198 & 7.196644400 & 17.079 & 40.841 & 22.531776 & 45.060293  \\
\texttt{book} & 500 & 2 & 880 & 20000 & 22401 & 5 & 134.476 & 311.635 & 1.596517534 & 3.706472180 & 9.806 & 22.328 & 20.906229 & 41.796343  \\
\texttt{c20ng} & 910 & 2 & 880 & 36400 & 38801 & 5 & 217.939 & 523.526 & 2.587875051 & 6.029774956 & 16.381 & 37.619 & 37.712999 & 75.396714  \\
\texttt{cr52} & 889 & 10 & 1768 & 35560 & 41985 & 7 & 304.390 & 700.521 & 3.619301009 & 8.340147404 & 21.114 & 49.053 & 38.085061 & 76.141513  \\
\texttt{cwebkb} & 839 & 10 & 1768 & 33560 & 39985 & 7 & 294.29858 & 676.65497 & 3.4984 & 8.0568 & 20.6358 & 46.8293 & 36.035452 & 72.043907  \\
\texttt{dna} & 180 & 22 & 4420 & 18000 & 34061 & 7 & 400.175 & 906.7995 & 4.76194 & 10.80538 & 24.4088 & 59.44160 & 22.5445 & 45.083  \\
\texttt{jester} & 100 & 2 & 4400 & 20000 & 32001 & 5 & 264.355 & 604.262 & 3.14534 & 7.1966 & 17.10386 & 40.8777 & 22.5317 & 45.0602  \\
\texttt{kdd} & 64 & 10 & 1768 & 2560 & 8985 & 7 & 136.70284 & 307.8906 & 1.6245 & 3.6652 & 8.5308 & 20.8365 & 4.266557 & 8.531005\\
\texttt{kosarek} & 190 & 2 & 2200 & 19000 & 25001 & 5 & 178.182 & 409.5368 & 2.11687 & 4.87362 & 12.3329875 & 28.7319 & 20.4866 & 40.967  \\
\texttt{msnbc} & 17 & 10 & 1768 & 680 & 7105 & 7 & 126.9435 & 285.3938 & 1.5108 & 3.3989 & 7.96834 & 19.07186 & 2.339926 & 4.679258\\
\texttt{msweb} & 294 & 22 & 4420 & 29400 & 45461 & 7 & 457.6946 & 1042.26683 & 5.451 & 12.420 & 29.0565 & 68.9942 & 34.221849 & 68.434206\\
%\texttt{nltcs} & 16 & 202 & 9800 & 3200 & 102201\\
\texttt{plants} & 69 & 2 & 4400 & 13800 & 25801 & 5 & 233.0806 & 530.5189 & 2.770567 & 6.318326 & 14.73527 & 35.330077 & 16.181983 & 32.361699 \\
\texttt{pumsb\_star} & 163 & 2 & 4400 & 32600 & 44601 & 5 & 328.2498 & 754.047 & 3.9069 & 8.9816 & 22.075988 & 52.048611 & 35.436199 & 70.867117\\
\texttt{tmovie} & 500 & 10 & 1768 & 20000 & 26425 & 7 & 225.16933 & 515.2254 & 2.6787 & 6.1358 & 14.937111 & 35.160811 & 22.139125 & 44.262135\\
\texttt{tretail} & 135 & 2 & 4400 & 27000 & 39001 & 5 & 299.88872 & 687.527 & 3.56848 & 8.1882965 & 19.84694 & 47.23953 & 29.7009 & 59.3974\\
\midrule
\multirow{5}{*}{$\texttt{nltcs}$} & 16 & 2 & 880 & 640 & 3041 & 5 & 36.0104 & 81.1858 & 0.4262 & 0.9636 & 2.6949 & 5.8308 & 1.066 & 2.1315\\
 & 16 & 2 & 2200 & 1600 & 7601 & 5 & 89.586 & 202.298 & 1.065 & 2.4087 & 5.857 & 13.845 & 2.664 & 5.426\\
 & 16 & 2 & 4400 & 3200 & 15201 & 5 & 178.932 & 404.3435 & 2.1298 & 4.8166 & 11.312 & 27.351 & 5.3258 & 10.651193\\
 & 16 & 10 & 1768 & 640 & 7065 & 7 & 126.802 & 284.8314 & 1.50844 & 3.39321 & 7.9706 & 19.55636 & 2.298934 & 4.597\\
& 16 & 22 & 4420 & 1600 & 17661 & 7 & 316.42786 & 711.7047 & 3.7706 & 8.4821 & 19.258 & 47.6545 & 5.746 & 11.491\\
\midrule
\multirow{4}{*}{$\texttt{nips}$} & {\multirow{2}{*}{100}} & {\multirow{2}{*}{7}} & {\multirow{2}{*}{17}} & {\multirow{2}{*}{1061}} & {\multirow{2}{*}{1084}} & {\multirow{2}{*}{11}} & 26.04271 & 71.28372 & 0.2977530 & 0.83463 & 2.06278 & 5.0131688 & 1.300823 & 2.601594\\
& & & & & & & 28.54619$^\dagger$ & 76.21134$^\dagger$ & 0.327307$^\dagger$ & 0.89374$^\dagger$ & 2.16724$^\dagger$ & 5.252681$^\dagger$ & 1.762611$^\dagger$ & 3.063386$^\dagger$ \\\cmidrule{2-15}
 & {\multirow{2}{*}{100}} & {\multirow{2}{*}{15}} & {\multirow{2}{*}{43}} & {\multirow{2}{*}{2750}} & {\multirow{2}{*}{2807}} & {\multirow{2}{*}{15}} & 65.803 & 182.21 & 0.77022 & 2.160 & 4.55309 & 12.382 & 3.043669 & 6.087289 \\
& & & & & & & 72.702458$^\dagger$ & 196.289$^\dagger$ & 0.854092$^\dagger$ & 2.32806$^\dagger$ & 4.822130$^\dagger$ & 12.872$^\dagger$  & 4.354196$^\dagger$ & 7.397817$^\dagger$\\
\bottomrule
\end{tabular}
\npnoround
%}
\label{table:benchmarks}
\end{table*}

Our benchmarks are given in Table~\ref{table:benchmarks}.
Compared to previous works \CHANGED{focused on NNs}, we evaluate a variety of datasets, which shows that \coolname{} can easily transform any SPN into a privacy-preserving version.
In addition to the theoretical analysis of Section~\ref{subsec:framework_efficiency}, we also investigate RAT-SPNs of various sizes for the~\texttt{nltcs} dataset of~\cite{gens2013learning} to 
gain a practical sense of how
different SPN parameters affect our runtime.
Moreover, we use two 
\emph{regular} SPNs trained for~\texttt{nips} to see how hiding the scope~(cf.~Section~\ref{subsubsec:privspn_scope}) increases the runtime.

Generally, our results shown in Table~\ref{table:benchmarks} demonstrate that we achieve tractable setup and highly efficient online performance for medium-sized~SPNs.
Specifically, the setup phase requires costs in the order of minutes and gigabytes, while the online phase takes only a few seconds and megabytes.
Though multiple seconds might seem like a significant slow-down in some cases, this is certainly justified in many scenarios where privacy demands outweigh the costs of privacy protection (such as legal requirements for 
medical diagnostics).

While no single parameter appears to be decisive for the runtimes, we observe that some parameters are much more significant:
\begin{enumerate}
    \item The number of sums has a significantly larger effect than products or leaves, which is expected given the log2 and exp2 operations. But, since the absolute amount of sums is still relatively small, the additional input weights do not affect online communication.
    \item Though differences in the number of~RVs, product nodes, leaves, and edges do influence the runtimes, deviations have to be very large to take an effect. For instance, when examining the SPNs for~\texttt{accidents}, \texttt{baudio}, and~\texttt{msweb}, it takes roughly twice the amount of RVs and edges (the SPN for~\texttt{msweb}) compared to the others to reach a significant runtime deviation.
    \item When looking at the SPNs for~\texttt{nltcs}, the first three SPNs have roughly the same density and the runtime seems to scale according to their size. The last two SPNs, however, have a noticeably higher density but comparable size and result in much higher runtimes. Thus, density (especially the amount of edges) is a much more significant parameter than plain network size. 
\end{enumerate}
Yet, depending on the SPN, the costs of other, less important parameters can outweigh the costs of individual parameters.
This is in line with our theoretical analysis in Section~\ref{subsec:framework_efficiency}: the circuit's size depends on the number of children (with different costs for sums and products) as well as the number of RVs and leaves. The amount of layers has no direct effect because the round complexity of Yao's~GC protocol is independent of the depth.
As for the regular~SPNs for~\texttt{nips}, one can observe that the effects of hiding~RV assignments 
are insignificant compared to the overall performance. 

Using~$64$-bit precision roughly doubles the costs of~$32$-bit precision, which is expected as the sub-circuits are about twice the size~\cite{demmler_automated_2015}.
Comparing the difference of the resulting \CHANGED{log-}probabilities when evaluating the~SPNs in~\coolname{} to the plain evaluation with~SPFlow, we get an~RMSE of~\num{4.2e-9} for~$32$-bit and~\num{2.3e-17} for~$64$-bit models.
We stress that this insignificant loss in accuracy is not due to the cryptographic measures, but rather due to the more~SMPC-friendly computation in the~log2 domain.

\section{CONCLUSION}
\label{sec_conclusion}
Resolving privacy issues in~ML applications is becoming a challenging duty for researchers, not least due to recent legal regulations such as the~GDPR.
By combining efforts from both~AI and applied cryptography research, we presented~\coolname{}, which successfully addresses this challenge for the evaluation of sum-product networks~(SPNs) that support a wide variety of desired~ML tasks.
The protocols of \coolname{} together with the tools developed for~ML experts deliver efficient yet extremely accurate~SPN inference while providing unprecedented protection guarantees that even cover the network scope and structure.
With our work serving as a foundation, future research can investigate further efficiency improvements (e.g., via quantization techniques appropriate for SPNs), hiding the structure of~SPNs that cannot be re-trained, and private~SPN learning.

\ack This project has received funding from the European Research Council~(ERC) under the~European Union's Horizon~2020 research and innovation programme~(grant agreement No.~850990 PSOTI). It was co-funded by the Deutsche Forschungsgemeinschaft~(DFG)~–- SFB~1119 CROSSING/236615297 and GRK~2050 Privacy \& Trust/251805230, and by the~BMBF and~HMWK within~ATHENE. KK also acknowledges the support of the Federal Ministry of Education and Research (BMBF), grant number 01IS18043B ``MADESI''.

\bibliography{bibliography}
\end{document}
%%%%%%%%%%%%%%%%%%%%%%%%%%%%%%%%%%%%%%%%%%%%%%%%%%%%%%%%%%%%%%%%%%%%%%